\documentclass{article}

\usepackage[preprint, nonatbib]{nips_2018}

\usepackage{float}

\usepackage[square,numbers]{natbib}
\usepackage[utf8]{inputenc} % allow utf-8 input
\usepackage[T1]{fontenc}    % use 8-bit T1 fonts
\usepackage{url}            % simple URL typesetting
\usepackage{booktabs}       % professional-quality tables
\usepackage{amsfonts}       % blackboard math symbols
\usepackage{nicefrac}       % compact symbols for 1/2, etc.
\usepackage{microtype}      % microtypography
\usepackage{graphicx}
\usepackage{tabularx}

\usepackage{amsmath}
\usepackage{algorithm}  
\usepackage{algpseudocode} 
\usepackage{subfigure}
\usepackage{multirow}

\title{GraphCL-DTA: a graph contrastive learning with molecular semantics for drug-target binding affinity prediction}

\author{
	Xinxing Yang \\
	Ningbo Artificial Intelligence Institute, Shanghai Jiao Tong University\\
	Department of Automation, Shanghai Jiao Tong University\\
	\texttt{yangxinxing@sjtu.edu.cn} \\
	\And
	Genke Yang \\
	Ningbo Artificial Intelligence Institute, Shanghai Jiao Tong University\\
	Department of Automation, Shanghai Jiao Tong University\\
	\texttt{gkyang@sjtu.edu.cn} \\
	\And
	Jian Chu \\
	Ningbo Artificial Intelligence Institute, Shanghai Jiao Tong University\\
	Department of Automation, Shanghai Jiao Tong University\\
	\texttt{chujian@sjtu.edu.cn} \\
}

\begin{document}
\maketitle

\begin{abstract}
	
Drug-target binding affinity prediction plays an important role in the early stages of drug discovery, which can infer the strength of interactions between new drugs and new targets. However, the performance of previous computational models is limited by the following drawbacks. The learning of drug representation relies only on supervised data, without taking into account the information contained in the molecular graph itself. Moreover, most previous studies tended to design complicated representation learning module, while uniformity, which is used to measure representation quality, is ignored. In this study, we propose GraphCL-DTA, a graph contrastive learning with molecular semantics for drug-target binding affinity prediction. In GraphCL-DTA, we design a graph contrastive learning framework for molecular graphs to learn drug representations, so that the semantics of molecular graphs are preserved. Through this graph contrastive framework, a more essential and effective drug representation can be learned without additional supervised data. Next, we design a new loss function that can be directly used to smoothly adjust the uniformity of drug and target representations. By directly optimizing the uniformity of representations, the representation quality of drugs and targets can be improved. The effectiveness of the above innovative elements is verified on two real datasets, KIBA and Davis. The excellent performance of GraphCL-DTA on the above datasets suggests its superiority to the state-of-the-art model.
	
\end{abstract}

\section{Introduction}

The process of developing a new drug for a specific disease requires an estimated cost of 260 million US dollars and about 17 years to pass the review of the Food and Drug Administration (FDA) \cite{1,2}. Small molecule drugs, as compounds, can cure related diseases by binding to specific targets. Therefore, the prediction of binding affinity between drugs and targets is a key task, which can accelerate drug discovery \cite{3}. 

Previous studies used high-throughput screening techniques to study the binding affinity between drugs and targets \cite{4}. This exhaustive method is not only expensive but also time-consuming. Moreover, this method lacks feasibility theoretically as there are millions of compounds currently known to humans and hundreds of targets that have therapeutic effects on diseases, the permutation and combination of which is excessive. Therefore, the development of high-precision computational models \cite{5} for predicting the binding affinity between drugs and targets can reduce the cost of identifying interactions between drugs and targets, and increase the speed as well, which is of great significance for promoting drug discovery.

The computational models \cite{6,widedta,am1,am2,am3} use the information of the drug and the target as the input, and the half maximal inhibitory concentration (IC50) after the drug binds to the target as the output. The binding affinity between the drug and the target reflects the interaction strength between them, which can provide more information on the relationship between the drug and the target. Computational models generally consist of three modules: drug representation learning module, target representation learning module, and prediction module. For example, Ozturk et al. \cite{m1} proposed the DeepDTA model, using the sequence information of drugs and proteins to predict the binding affinity between them. DeepDTA uses the convolutional neural network (CNN) as representation learning module for drugs and targets. They used simplified molecular input line entry system (SMILES) strings and protein sequences as raw input. Then the representations of the drug and target are concatenated into the fully connected neural network to obtain the predicted binding affinity. On the basis of DeepDTA, Abbasi et al. \cite{m2} proposed the DeepCDA model. The representation learning module of this model combines CNN and long short-term memory network (LSTM). The workflow is to firstly input the sequence information of drug and target into CNN and LSTM sequentially to capture the global information among the sequence information. Then an attention mechanism was used in their prediction module, which is able to infer the importance weights in sequences. Similarly, Zeng et al. \cite{m3} proposed the MATT-DTI model, using the SMILES strings as the input of the drug representation learning module and the protein sequence as the input of the target representation learning module. The prediction module adopts multi-head attention mechanism and multi-layer neural network. In the drug representation learning module, they proposed a relation-aware self-attention model for learning drug representations, which could capture the correlation of atoms in drugs. In the target representation learning module, they employed a one-dimensional convolutional neural networks (1D CNN) to learn the representation of the target. Then the drug representation and the target representation were concatenated and input into the prediction module to obtain the value of the predicted binding affinity. 

Representing drugs with SMILES strings is not a natural way. Therefore, Nguyen et al. \cite{m4} proposed the GraphDTA model and designed a drug representation learning module based on molecular graphs. The drug representation learning module uses the graph Convolutional network (GCN) to learn the structural information from the molecular graph, and then obtains the drug representation through the global maximum pooling operation. Likewise, this model learns object representations by use of protein sequences and 1D CNN. In the prediction module, they concatenated the drug representation and the target representation and input it into the fully connected neural network to obtain the predicted binding affinity. Validated on multiple datasets, drug representations learned by using molecular graphs have better generalization ability than SMILES strings. Bidirectional LSTM has demonstrated its superiority over 1D CNN in capturing the characteristics of temporal signals in many fields. Therefore, on the basis of GraphDTA, Mukherjee et al. \cite{m5} proposed a new drug-target binding affinity prediction model. The difference is that this model uses a bidirectional LSTM model as the target representation learning module. The experimental results showed that the performance of the bidirectional LSTM model is better than that of 1D-CNN. Li et al. \cite{m6} improved the prediction module and proposed a two-way attention neural network for the fusion of drug and target representations. This approach not only enables the model to focus on the effective sites of nuclear amino acids, but also makes itself interpretable.

However, the previous work has the following two limitations. Firstly, the above methods input the molecular graph into the GCN \cite{gcn1,gcn2}, and uses the supervised data of known drug-target binding affinity to inversely optimize the parameters in the GCN to learn the drug representation. The disadvantage is that it relys on a large amount of supervised data to learn the appropriate drug representation, which is often time-consuming and expensive. Moreover, such drug representation is dependent on the supervised data. Without using the data of the molecular graph itself, the characteristics of the drug cannot be described in essence.

Graph contrastive learning \cite{gcl}, as a self-supervised framework, is able to learn high-level intrinsic features from the drug molecular graph itself to distinguish different drugs without supervised data. However, previous data augmentation strategies mainly use random deletion of edges or nodes to generate different views, which changes the semantics of molecular graphs and cannot benefit from the relational inductive bias of molecular graphs. As shown in Figure 1, the deletion of nodes and edges on the molecular graph of aspirin produces molecules with other properties \cite{example}.

\begin{figure}[h]
	\centering
	\includegraphics[scale=0.35]{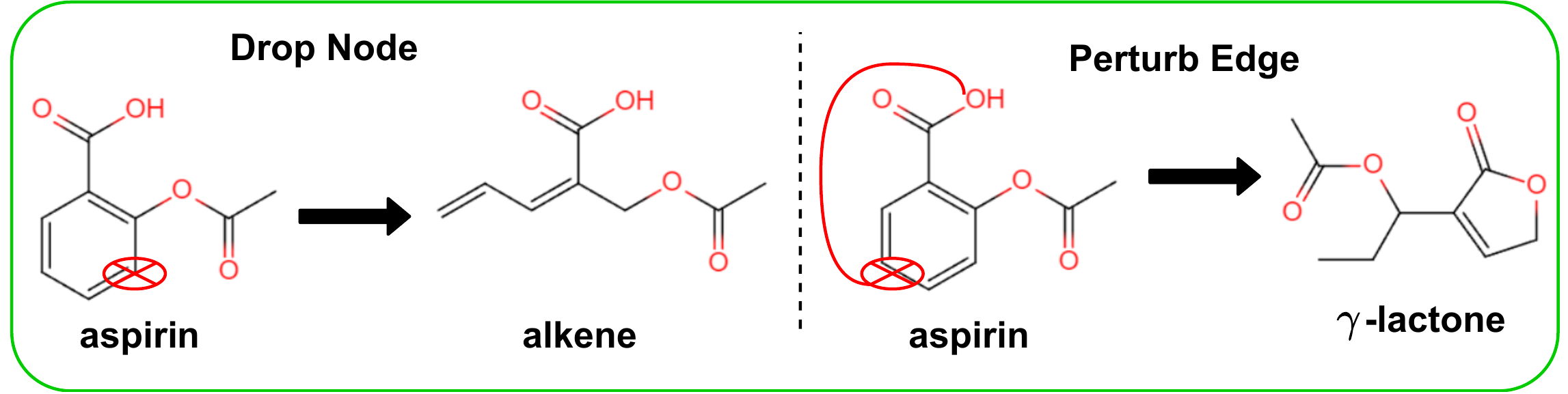}
	\caption{Dropout-based augmentation strategies destroy molecular semantics.}
	\label{}
\end{figure} 

Secondly, the above models use Mean Square Error (MSE) as the loss function for parameter optimization. This loss function does not take into account the uniformity of drug representation and target representation. At the same time, in the field of recommender systems, studies have shown that the uniformity of representation can improve the performance of the model \cite{cl1,cl2}. Previous computational models focused on how to design effective representation learning modules and prediction modules, but few studies have been conducted to improve the loss function of drug-target binding affinity.

Therefore, in response to the above limitations, inspired by graph contrastive learning \cite{simcl} and representation uniformity \cite{unio}, we propose GraphCL-DTA, a graph contrastive learning with molecular semantics for drug-target binding affinity prediction. The GraphCL-DTA model consists of a drug representation learning module, a target representation learning module and a prediction module. In the drug representation learning module, we propose a graph contrastive learning framework that preserves molecular semantics. The framework flow is as follows: 1. Input the molecular graph of the drug into the GCN to obtain the drug representation. 2. Add different random noises to the drug representation to generate two contrastive views. 3. Optimize the contrastive views so that the views belonging to the same drug become closer in the representation space, and views belonging to different drugs become farther apart. The advantage of this framework is that complex data augmentation strategies are no longer needed to generate contrastive views, and contrastive views can be created by adding controllable random noise directly in the representation space of drugs, resulting in variance across different contrasted views. By controlling the magnitude of random noises, the semantics of molecular graphs are preserved. Through this graph contrastive framework, a more essential and effective drug representation can be learned without additional supervised data.

In the target representation, we use the same operation as GraphDTA, which is to input the protein sequence into the 1D CNN to obtain the target representation. In the prediction module, we concatenate the drug representation and the target representation to form a one-dimensional vector, which is then input into a multi-layer fully connected neural network to obtain the predicted binding affinity.

In addition, we design a new loss function that can be directly used to smoothly adjust the uniformity of drug representation and target representation. Based on MSE, the loss function adds a regular term to directly optimize the uniformity of the representation. By minimizing this loss function, the uniformity and quality of drug representation and target representation are enhanced, thereby improving the predictive ability of the GraphDTA model.

In general, the main contributions of this work are as follows:

\begin{enumerate}
	\item We design a graph contrastive learning framework for learning drug representations. Semantic information of molecular graphs can be preserved without data augmentation strategies. Through this graph contrastive framework, a more essential and effective drug representation can be learned without additional supervised data.
	\item We design a loss function optimizing the uniformity of drug and target representations to improve the representation quality.
	\item  Extensive experiments have been conducted on two publicly available datasets. The results demonstrate the effectiveness of both the graph contrastive learning framework and the loss function optimized for uniformity. The performance of the GraphCL-DTA model is also verified to be superior over the state-of-the-art model.
\end{enumerate}

The rest of the paper is organized as follows. Section 2 introduces our proposed GraphCL-DTA model. Section 3 introduces the relevant experimental results. Section 4 makes the conclusion.

\begin{figure}[h]
	\centering
	\includegraphics[scale=0.5]{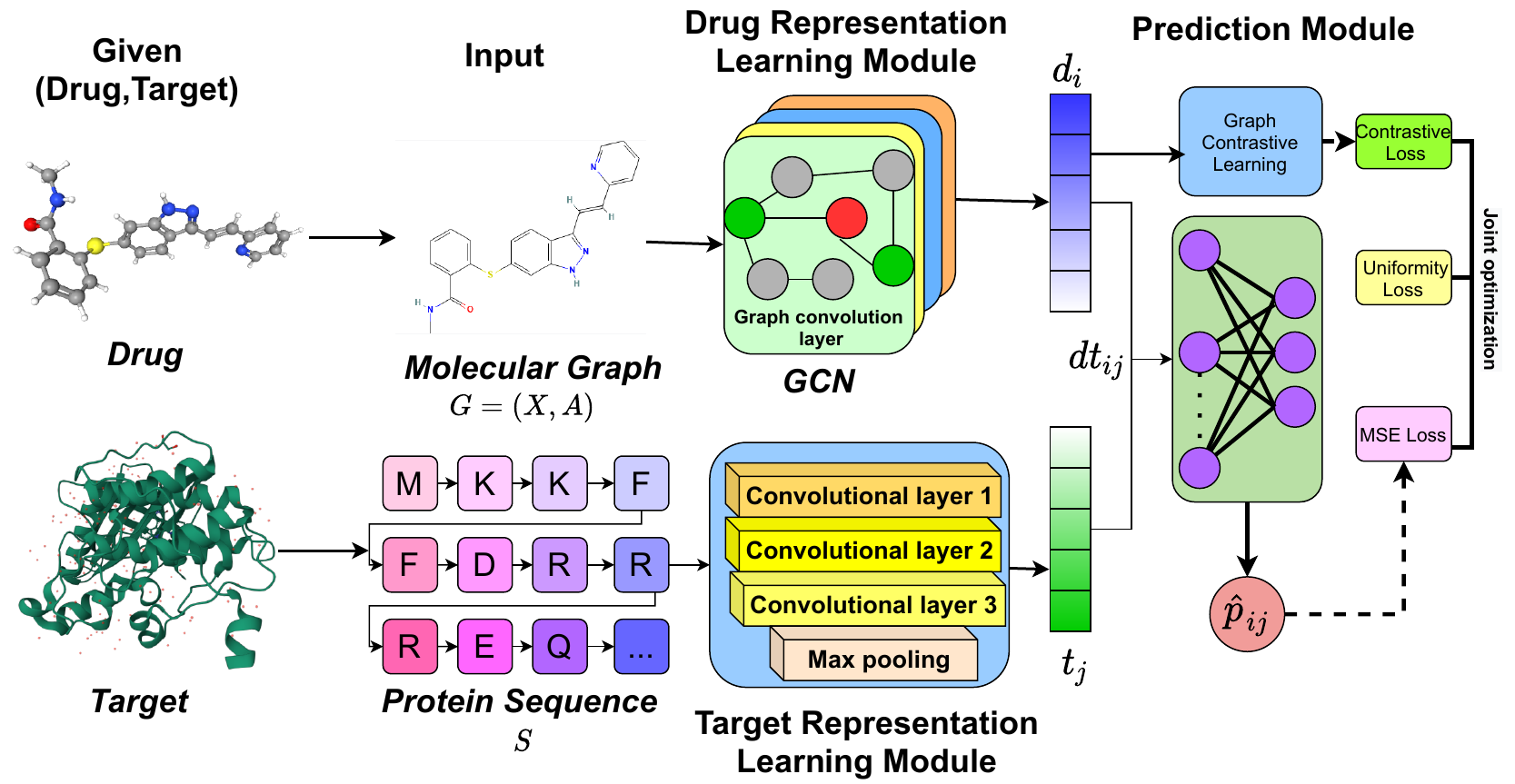}
	\caption{The architecture of our proposed GraphCL-DTA model.}
	\label{}
\end{figure} 

\section{Method}

Figure 2 shows the architecture of our proposed GraphCL-DTA model. Obviously, the GraphCL-DTA model consists of three modules, which are the drug representation learning module, the target representation learning module, and the prediction module. The workflow of the GraphCL-DTA model, simply put, is that the molecular graph of the drug is input into the drug representation learning module to obtain the representation of the drug, and the protein sequence of the target is input into the target representation learning module to obtain the representation of the target. The representation of the drug and the target are concatenated and then input into the prediction module to obtain the predicted binding affinity.

The drug representation learning module is the focus of this work, in which we design a graph contrastive learning framework for molecular graphs, which can learn better drug representation without complicated manual augmentation strategies and additional supervised data. The target representation learning module is similar to GraphDTA, both of which use a 1D CNN. The prediction module uses a fully connected neural network. Another focus of this work is a new loss function for directly optimizing the uniformity of the representation, thereby improving the quality of the representation of drugs and targets. Firstly, we illustrate how to use GCN to learn drug representations. Then we sequentially introduce the target representation, prediction module and optimization function. Next, we explain how to use the graph contrastive framework to improve the representation quality of drugs. The final part is the loss function optimized for uniformity.

\subsection{Drug Representation using GCN}

Given a drug $i$, its molecular graph can be defined as $G=(X,A)$. $X\in \mathbb{R}^{n\times k}$ represents the features matrix of atoms. $n$ represents the number of atoms. $k$ represents the feature number of each atom. $A\in \mathbb{R}^{n\times n}$ represents the adjacency matrix. If $A[i][j]=1$, it means that there is a bond connection between atom $i$ and atom $j$. Otherwise there is no bond connection. The GCN takes $G=(X,A)$ as input and outputs $d_i \in \mathbb{R}^{1\times h}$, representing the representation of drug $i$. The output $Z^l$ of $l$th-layer in GCN is defined as follows. The propagation rules of each layer in GCN are shown in equation (1).

\begin{equation}
Z^{l+1} = \sigma(\tilde{D}^{-\frac{1}{2}} \tilde{A} \tilde{D}^{-\frac{1}{2}} Z^{l} W^{l})
\end{equation}

$\tilde{A}=A+I$, $I$ represents the identity matrix. The purpose of this operation is to add a self-connection to the adjacency matrix. $\tilde{D}_{ii}=\sum_{j}\tilde{A}_{ij}$. $Z^{l} \in \mathbb{R}^{n\times h}$ represents the input of the $l$th layer. Note that $Z^{0}=X$. $W^l$ represents the trainable weight matrix of the $l$ layer. We define the output of the last layer of the GCN as $Z^{F}$, which is the representation of the node-level. We need the representation of the graph-level, so we use the equation (2) for $Z^ {F}$ do global maximum pooling operation.

\begin{equation}
	d_i = GMP(Z^{F})
\end{equation}

Among them, $GMP$ represents the global max pooling layer, which removes the maximum value of each column of $Z^{F}$, forming a vector $d_i$ of shape $1\times h$, to represent drug $i$.

\subsection{Target Representation using 1D CNN}

We adopt the same approach as GraphDTA to learn the representation of targets. The protein sequence of the target is input into three 1D CNN layers and a maximum pooling layer to obtain the representation vector of the target. The protein sequence of the target uses a string of ASCII characters representing amino acids, and we use integer encoding to convert it into an integer sequence vector (Alanine (A) is 1, Cystine (C) is 3, etc.). Given a target $j$, its protein sequence is $S$. We input it into the target representation learning module to get its representation $t_j \in \mathbb{R}^{1\times h}$.

\subsection{Prediction Layer and optimization function}

In the prediction module, we concatenate $d_i$ and $t_j$ into a $1\times 2h$ vector $dt_{ij}$, and then input $dt_{ij}$ into a three-layer fully connected neural network to obtain the predicted drug-target binding affinity $\hat{p}_{ij}$. The propagation of the fully connected function is shown in the equation (3).

\begin{equation}
	 \hat{p}_{ij}=  V_3^T \sigma(V_2^T\sigma(V_1^Tdt_{ij}+b_1)+b_2)+b_3	
\end{equation}

$dt_{ij} \in\mathbb{R}^{1\times2 h}$, $V_1^T$ , $V_2^T$ and $V_3^T\in\mathbb{R}^{2h\times 1}$ are the weight parameters of each fully connected neural network. $b_1$, $b_2\in1 *h$ and $b_3$ are paranoid parameters; $\hat{p}_{ij}$ represents the binding affinity between drug $i$ and target $j$ predicted by the GraphCL-DTA model.

\begin{equation}
\mathcal{L}_{mse}	= \frac{1}{c} \sum_{1}^{c} (p_{ij} - \hat{p}_{ij})^2
\end{equation}

As shown in equation (4), we use MSE as the loss function to optimize the parameters in the model. By minimizing the $\mathcal{L}_{mse}$ in equation (4), the predicted value $\hat{p}_{ij}$ can continuously approach the real value $p_{ij}$, thereby generating a gradient, which is backpropagated to the GraphCL-DTA model for learning appropriate parameters.

\subsection{Graph contrastive learning for drug representations}

Previous dropout-based augmentation strategies change the semantics of molecular graphs, which is not suitable for drug scenarios. We therefore directly introduce the data augmentation strategy into the drug embedding space. This is inspired by adversarial training in the image domain and simple graph contrastive learning \cite{ade,simcl}, where adding controllable noise to the embedding space of an image does not change the semantic information. We directly add random noise to the drug embedding space as a simple and effective augmentation strategy. The data augmentation strategy is shown in equation (5).

\begin{equation}
	e_i^1 = d_i +\Delta^{'}_i, \quad e_i^2 = d_i +\Delta^{''}_i
\end{equation}
 
Random noise $\Delta^{'}_i $ and $\Delta^{''}_i $ are subject to two constraints. The first one is that $\parallel \Delta^{'}_i \parallel_2 = \parallel \Delta^{''}_i \parallel_2 = \epsilon $, and the second is that the $\Delta^{'}_i,\Delta^{''}_i = \tilde{\Delta}\odot sign(d_i)$, $ \tilde{\Delta} \in \mathbb{R}^{d}\sim U(0,1) $. The effect of the first constraint is to control the magnitude of the noise vector, which is numerically equivalent to a point on a hypersphere with radius $\epsilon$. The second restriction is set to ensure that the three vectors, $e_i^1 $ and $e_i^2 $, $d_i$ are in the same hyperoctant. In this way, the contrastive views, $e_i^1 $ and $e_i^2 $, will not have large semantic deviations from the original view $d_i$.

\begin{figure}[h]
	\centering
	\includegraphics[scale=0.5]{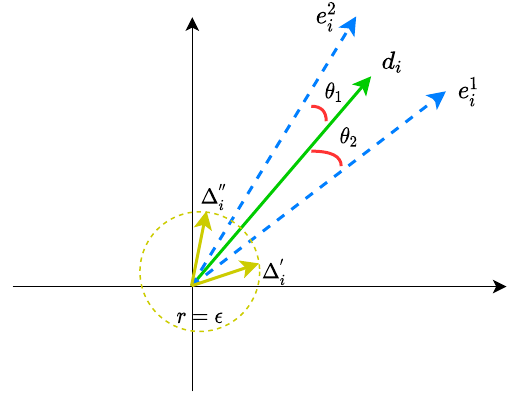}
	\caption{An example of a data augmentation strategy in $ \mathbb{R}^2$ Space.}
	\label{}
\end{figure} 

An example of a data augmentation strategy in $ \mathbb{R}^2$ Space is shown in Figure 3. The representation of the drug, $d_i$, is rotated by a small angle in the embedding space under the influence of noise $\theta_1$ and $\theta_2$. Each angle of rotation directly produces an enhanced representation ($e_i^1$ and $e_i^2$)that can be directly used as a contrastive view. It is worth noting that as long as the rotation angle is small enough, the contrastive view can not only retain the maximum effective information of the original representation, which means that the semantic information of the drug is not changed, but also cause variance between the two representations. This makes it easier for GCL to learn information. The contrastive losses are shown below equation (6).

\begin{equation}
\mathcal{L}_{gcl} = \sum_{i\in \mathcal{B}} -\log \frac{exp(e_i^1 e_i^2/\tau)}{\sum_{j\in \mathcal{B}} exp(e_i^1 e_j^2/\tau)}
\end{equation} 

Both drug $i$ and drug $j$ are obtained from the same batch $\mathcal{B}$. $e_i^1 $ and $e_i^2 $ represent two contrastive views of drug $i$. $e_j^2 $ represents the contrastive view of drug $j$. The effect of the contrastive loss is to strengthen the consistency of the two contrastive views belonging to the drug $i$, while the consistency of the contrastive views belonging to different drugs is weakened. To make graph contrastive learning effective, we jointly optimize the contrastive loss $\mathcal{L}_{gcl}$ and the MSE loss $\mathcal{L}_{mse}$, as shown in equation (7) below. $\alpha$ is used to adjust the weight of the latter.

\begin{equation}
	\mathcal{L}_{joint} = \mathcal{L}_{mse} + \alpha \mathcal{L}_{gcl}
\end{equation}

There are two reasons why the above graph contrastive learning can improve drug representation. Firstly, through the contrastive loss $\mathcal{L}_{gcl}$, the structural information can be learned from the molecular graph itself without supervised data, making each drug representation highly discriminative. Secondly, the contrastive loss $\mathcal{L}_{gcl}$ can be used as a regular term in joint optimization loss $\mathcal{L}_{joint}$ to limit the complexity of the GraphCL-DTA model, remove redundant information in the representation, and thus extract its generalization ability in subsequent prediction tasks.

\subsection{Loss function optimized for uniformity of representation}

Different from previous loss functions, we design a new version to directly optimize the uniformity of drug representation and target representation, which has been proven in the field of data mining to improve the generalization performance of the model \cite{unio}. Firstly we measure the uniformity of drug representation and target representation using equations (8) and (9).

\begin{equation}
	\mathcal{L}_{uniform}^{d} = \log \mathop{\mathbb{E}}\limits_{u,v\sim D} e^{-2\parallel d_u - d_v \parallel_2}
\end{equation}

\begin{equation}
\mathcal{L}_{uniform}^{t} = \log \mathop{\mathbb{E}}\limits_{u,v\sim T} e^{-2\parallel t_u - t_v \parallel_2}
\end{equation}

Taking the uniformity of drug representation as an example, we are actually computing the logarithm of the average pairwise Gaussian potential. Therefore, as shown in equation (10), we directly add uniformity to the loss function, making the model optimize it directly.  $\beta$ is used to adjust the weight of the $\mathcal{L}_{uniform}^{d}$ and $\mathcal{L}_{uniform}^{t}$.

\begin{equation}
\mathcal{L}_{joint} = \mathcal{L}_{mse} + \alpha \mathcal{L}_{gcl} + \beta (\mathcal{L}_{uniform}^{d}+\mathcal{L}_{uniform}^{t})
\end{equation}

The uniformity of the representation is helpful to the GraphCL-DTA model. Firstly, it can be used as a regular term in the loss function as well to improve the generalization performance of the model. Secondly, the uniform performance of representations improves the model as the low similarity between the representations of different drugs makes each representation more discriminative.

\section{Experiments and Discussion}

\subsection{Datasets}

\begin{table}[h]
	\caption{The statistics of datasets}
	\centering
	\setlength{\tabcolsep}{5mm}
	\begin{tabular}{@{}ccc@{}}
		\toprule
		& Davis & KIBA   \\ \midrule
		Targets       & 442   & 229    \\
		Drugs         & 68    & 2111   \\
		Interactions  & 30056 & 118254 \\
		Training samples & 25046 & 98545  \\
		Test samples     & 5010  & 19709  \\ \bottomrule
	\end{tabular}
\end{table}

In this section, we use two mainstream datasets, Davis \cite{davis} and KIBA \cite{KIBA}, to evaluate the proposed model. As shown in Table 1, the Davis dataset contains 442 proteins, 68 drugs and 30,056 drug-target binding affinity values, where the values range from 5 to 10.8. The KIBA dataset contains 229 proteins, 2111 drugs and 118254 drug-target binding affinity values, where the values range from 0 to 17.2.

\subsection{Experimental settings and evaluation metrics}

Consistent with the settings of models such as GraphDTA, we divide all drug-target binding affinity values in the dataset into 6 equal parts, one of which is used as the test set. The rest uses 5-fold cross-validation to train the model and determine hyperparameters. We use three mainstream evaluation metrics to measure model performance, namely Mean Square Error (MSE), Concordance Index (CI) and $r^2_m$. The smaller the value of MSE, the better the performance of the model. The larger the value of CI and $r^2_m$, the better the performance of the model. Equations (10)-(13) show the calculation of these metrics.

\begin{equation}
	MSE = \frac{1}{N} \sum_{i}(p_i- \varrho_i)^2
\end{equation}

\begin{equation}
CI=\frac{1}{T}\sum_{\varrho_i > \varrho_j}\mathcal{H}(p_i-p_j)
\end{equation}

\begin{equation}
\mathcal{H}(x)=\left\{
\begin{aligned}
1,  &  \quad x>0 \\
0.5,   &  \quad x=0 \\
0,  &  \quad x<0
\end{aligned}
\right.
\end{equation}

\begin{equation}
r_m^2 = r^2\times (1-\sqrt{r^2-r^2_0})
\end{equation}
$p$ is the prediction result corresponding to binding affinity $\varrho $. $T$ is a regularization term. $\mathcal{H}$ is a piecewise function. $r^2$ represents the square correlation coefficient, and the difference between $r^2_0$ and the former is that its intercept is 0.

\begin{table}[h]
	\caption{The setting of hyperparameters}
	\centering
	\setlength{\tabcolsep}{8mm}
	\begin{tabular}{@{}ccc@{}}
		\toprule
		Hyperparameters                  & Change interval               & Default value \\ \midrule
		Learning rate                    & $[0.0001,0.0005,0.001,0.005]$ & 0.0005        \\
		Batch size                       & $[128,256,512.1024]$          & 512           \\
		GCN layers                       & $[1,2,3,4,5]$                 & 3             \\
		Drug representation dimension    & $[64,128,256,512]$            & 128           \\
		Target representation dimension  & $[64,128,256,512]$            & 128           \\
		Contrastive loss weight $\alpha$ & $[0, 0.01, 0.1, 0.5, 1]$      & 0.5           \\
		Uniformity loss weight $\beta$   & $[0, 0.01, 0.1, 0.5, 1]$      & 0.5           \\ \bottomrule
	\end{tabular}
\end{table}

\subsection{Hyperparameter settings}

Table 2 shows the hyperparameters and their default values in the GraphCL-DTA model. The specific settings are as follows. The change interval of the learning rate $lr$ is $[0.001,0.005,0.01,0.05]$, the change interval of the batch size of the training set and the test set is $[128,256,512.1024]$. The number of graph convolutional network layers changes in the interval $[1,2,3,4,5]$. The change intervals of the drug representation dimension and the target point representation dimension are both $[64,128,256,512]$. The change intervals of the contrastive loss function weight and the uniformity loss weight are both $[0, 0.01, 0.1, 0.5, 1]$. Appropriate hyperparameter values are determined from these intervals through the validation set. The default values of the above hyperparameters are 0.005, 512, 3, 128, 128, 0.5 and 0.5.

\subsection{Ablation experiment}

In this section, we use ablation experiments to verify the effectiveness of the innovative elements proposed in this work, that is, whether the graph contrastive learning framework for drugs can improve the performance of the model, and whether the loss function optimized for representation uniformity is helpful to the model.

\begin{table}[h]
	\caption{The experimental results of the GCN-DTA, GCL-DTA and GraphCL-DTA on the Davis and KIBA datasets}
		\centering
	\setlength{\tabcolsep}{10mm}
	\begin{tabular}{@{}ccccc@{}}
		\toprule
		Dataset & Methods & MSE   & CI    & $r^2_m$ \\ \midrule
		Davis   & GCN-DTA     & 0.293 & 0.872 & 0.636   \\
		& GCL-DTA     & 0.259 & 0.884 & 0.648   \\
		& GraphCL-DTA     & 0.249 & 0.888 & 0.665   \\ \midrule
		KIBA    & GCN-DTA      & 0.140 & 0.887 & 0.767   \\
		& GCL-DTA      & 0.141 & 0.881 & 0.769   \\
		& GraphCL-DTA      & 0.138 & 0.888 & 0.778   \\ \bottomrule
	\end{tabular}
\end{table}

\subsubsection{Effectiveness of Graph Contrastive Learning}
In this work, we design a graph contrastive learning framework for molecular graphs to learn drug representations. By controlling the magnitude of random noise, the semantics of molecular graphs are preserved. The graph contrastive framework is able to learn more essential and effective drug representations without additional supervised data. To show whether this graph contrastive learning framework improves the model's performance, we compare it to the model described below. The first comparison model is GCN-DTA. This model is equivalent to the GraphCL-DTA model without the graph contrastive learning framework, which directly uses the GCN to learn the representation of drugs. The second comparison model is GCL-DAT, which replaces the graph contrastive learning in the GraphCL-DTA model with the dropout-based graph contrastive learning. The purpose of this comparison is to observe the impact of the semantics of molecular graphs on model performance.

The experimental results of the above models on the Davis and KIBA datasets are shown in Table 3. Firstly, by comparing the GraphCL-DTA model with the GCN-DTA model, we find that the performance of the GraphCL-DTA model has been improved with the support of the graph contrastive learning framework. Specifically, on the David dataset, the values of the three metrics of the GraphCL-DTA model are 0.249, 0.888 and 0.665, while on the KIBA dataset, the values are 0.138, 0.888 and 0.778. Compared with the GCN-DTA model, the three metrics of the GraphCL-DTA model on the David dataset have been improved by 15\%, 1.8\% and 4.5\% respectively. The three metrics on the KIBA dataset have increased by 1.4\%, 0.1\% and 1.4\% respectively. It is worth noting that the GraphCL-DTA model has an obvious effect on the Davis dataset, while showing a small improvement effect on the KIBA dataset. This is mainly because the Davis dataset has less supervised data, which becomes a hindrance to drug representation learning. Without relying on supervised data, the graph contrastive learning learns better drug representation by using drug molecular graphs for self-supervised learning, thereby improving model performance. The KIBA dataset has more supervised data, as a result of which, the improvement effect of the graph contrastive learning is limited. However, the upper limit of the performance of the model can still be improved to a certain extent.

In addition, the GCL-DTA model improves the performance of the GCN-DTA model on the data-sparse Davis dataset, but it does not perform well on the KIBA dataset with a huge amount of data. This is mainly due to the fact that the GCL-DTA model uses the dropout method to generate contrast views, which destroys the semantic information of molecules to a certain extent, so that the model learns wrong drug features.

Most importantly, compared with the GCL-DTA model, the three metrics of the GraphCL-DTA model on the David dataset have increased by 3.8\%, 0.4\% and 2.6\% respectively. The three metrics on the KIBA dataset have increased by 2.1\%, 0.7\% and 1.1\% respectively. This shows that compared with the dropout-based graph contrastive learning, the graph contrastive framework we propose retains the semantic information of molecule graph, and does not need to design complex data augmentation strategy, which has a positive effect on model performance. The discussion above validates the effectiveness of the graph contrastive learning framework proposed in this work for molecular graphs.

\begin{table}[h]
	\caption{The performance of the GraphCL-DTA model under different uniformity loss weights}
	\centering
	\setlength{\tabcolsep}{10mm}
	\begin{tabular}{@{}ccccc@{}}
		\toprule
		Dataset & Weight $\beta$     & MSE   & CI    & $r^2_m$ \\ \midrule
		Davis   & value=0    & 0.289 & 0.864 & 0.606   \\
		& value=0.01 & 0.278 & 0.869 & 0.649   \\
		& value=0.1  & 0.252 & 0.889 & 0.675   \\
		& value=0.5  & 0.261 & 0.882 & 0.667   \\
		& value=1    & 0.265 & 0.881 & 0.647   \\ \midrule
		KIBA    & value=0    & 0.138 & 0.890 & 0.783   \\
		& value=0.01 & 0.132 & 0.895 & 0.784   \\
		& value=0.1  & 0.133 & 0.895 & 0.789   \\
		& value=0.5  & 0.129 & 0.895 & 0.805   \\
		& value=1    & 0.132 & 0.893 & 0.791   \\ \bottomrule
	\end{tabular}
\end{table}

\begin{figure}[h]
	\centering
	
	\subfigure[MSE]{
		\includegraphics[width=4cm,height=4cm]{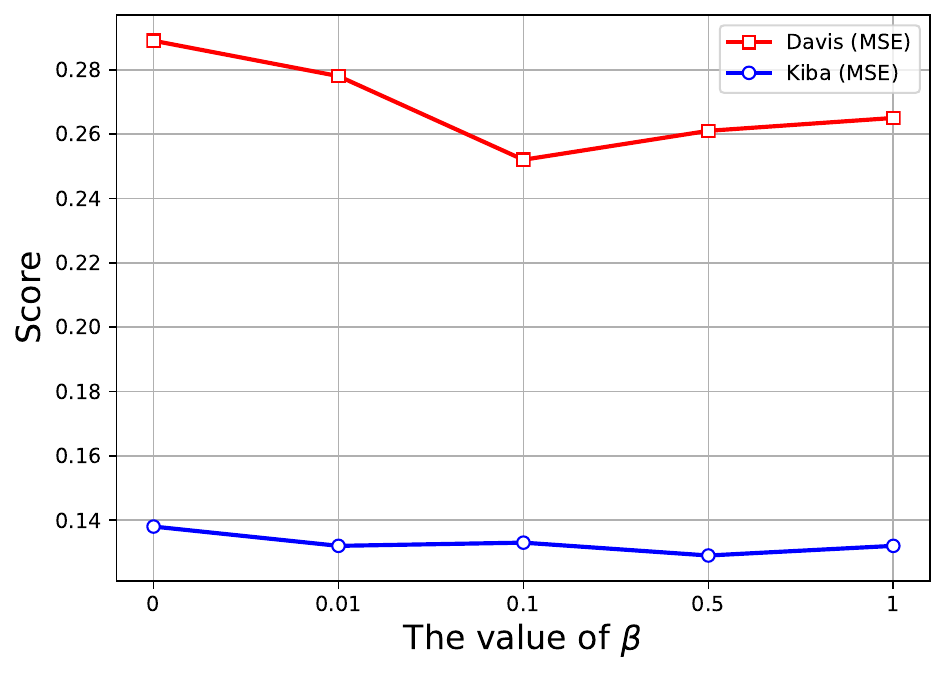}
		%\caption{fig1}
	}
	\quad
	\subfigure[CI]{
		\includegraphics[width=4cm,height=4cm]{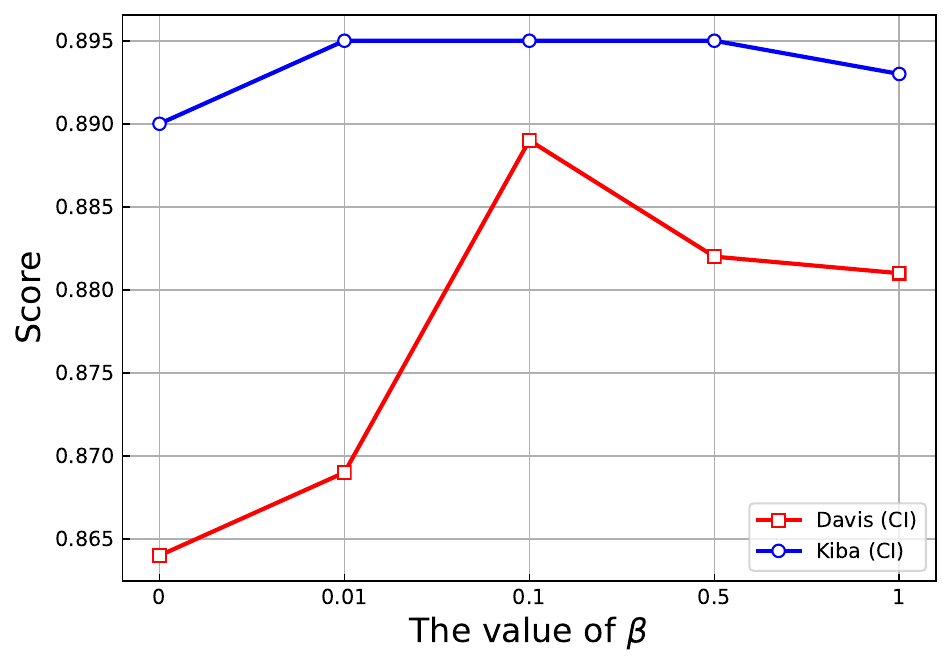}
	}
	\quad
	\subfigure[$r^2_m$]{
		\includegraphics[width=4cm,height=4cm]{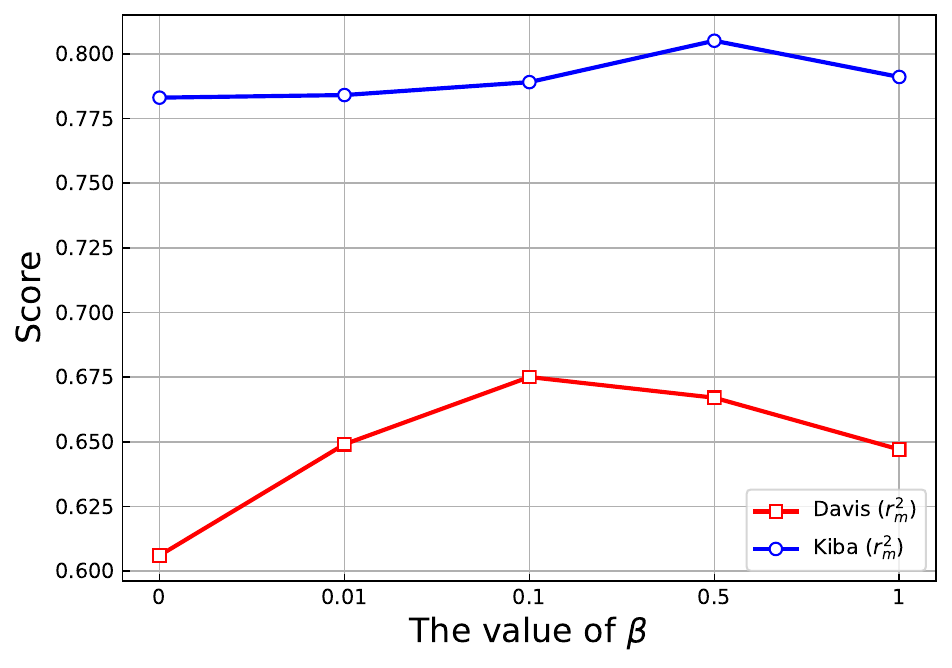}
	}
	
	\caption{ The sensitivity of the GraphCL-DTA model under different uniformity loss weights}
\end{figure}

\subsubsection{Effectiveness of Optimizing for Representation Uniformity}

In this work, we design a new loss function that can be directly used to smoothly adjust the uniformity of drug representation and target representation. Uniformity of representation is a property defined in the field of recommender systems to measure the quality of representation, and later found to be useful for improving model performance. To explore whether this loss function optimized for representation uniformity helps improve the performance of the GraphCL-DTA model, we run the following experiments. We change the value of the uniformity loss weight $\beta$ in the interval $[0, 0.01, 0.1, 0.5, 1]$ to observe the impact of different uniformity losses on the model performance.

Table 4 shows the performance of the GraphCL-DTA model under different uniformity loss weights $\beta$. Figure 4 shows the sensitivity of the GraphCL-DTA model under different uniformity loss weights. Firstly, when the value of the uniformity loss $\beta$ is equal to 0, that is, the uniformity is not optimized, the prediction performance of the GraphCL-DTA model on both the Davis and KIBA datasets is the worst. As long as the value of uniformity loss $\beta$ is greater than 0, its predictive performance on both data sets is better, thus verifying the importance of expressing uniformity. Specifically, when the value of $\beta$ is equal to 0, the performance of the GraphCL-DTA model on the Davis and KIBA datasets are $[0.289, 0.864, 0.606]$ and $[0.138, 0.890, 0.783]$, respectively. The best experimental results of the GraphCL-DTA model on the two datasets are $[0.252, 0.889, 0.675]$ and $[0.129, 0.895, 0.805]$, respectively, with an average increase of 9\% and 3.2\%. Secondly, the sensitivity of Davis and KIBA datasets to $\beta$ is different. On the Davis dataset, the GraphCL-DTA model achieves the best performance when the uniformity loss weight $\beta$ is 0.1. While on the KIBA dataset, the value of $\beta$ is 0.5. This may be caused by the different data scales of the two datasets.

Therefore, the above discussion shows that directly optimizing the uniformity of drug representation and target representation can distribute the representation of drugs or targets evenly on the sphere, improving the representation quality of drugs and targets, thereby enhancing the predictive ability of the GraphCL-DTA model.

\subsection{ Comparison with the state-of-the-art methods}

\begin{table}[h]
	\caption{The performance of the GraphCL-DTA model with state-of-the-art models}
	\centering
	\setlength{\tabcolsep}{6mm}
	\begin{tabular}{@{}ccccc@{}}
		\toprule
		Dataset & Methods    & MSE   & CI    & $r_m^2$ \\ \midrule
		Davis   & DeepDTA  (Öztürk et al., 2018)        & 0.261 & 0.878 & 0.630   \\
		& WideDTA    (Öztürk et al., 2018)        & 0.262 & 0.886 & 0.633   \\
		& DeepCDA       (Abbasi et al., 2020)    & 0.248 & 0.891 & 0.649   \\
		& GraphDTA    (Nguyen et al., 2021)        & 0.241 & 0.887 & 0.679   \\
		& DeepGLSTM    (Mukherjee et al., 2022)      & 0.236 & 0.893 & 0.677   \\
		& GraphCL-DTA      & 0.236 & 0.894 & 0.685   \\ \midrule
		KIBA    & DeepDTA    (Öztürk et al., 2018)       & 0.194 & 0.863 & 0.673   \\
		& WideDTA    (Öztürk et al., 2018)       & 0.179 & 0.875 & 0.675   \\
		& DeepCDA     (Abbasi et al., 2020)      & 0.176 & 0.889 & 0.682   \\
		& GraphDTA    (Nguyen et al., 2021)        & 0.151 & 0.883 & 0.687   \\
		& DeepGLSTM     (Mukherjee et al., 2022)     & 0.143 & 0.890 & 0.780   \\
		& GraphCL-DTA              & 0.129 & 0.895 & 0.805       \\ \bottomrule
	\end{tabular}
\end{table}

We compare the GraphCL-DTA model with the state-of-the-art methods in the field of drug-target binding affinity prediction to verify the superiority of the former. These state-of-the-art methods are described below.

\begin{itemize}
	\item \textbf{DeepDTA} \cite{m1}: DeepDTA inputs the SMILES string of the drug and the protein sequence of the target into the convolutional neural network to learn the representation of the drug and the target. Then DeepDTA inputs the concatenated drug representation and target representation into a fully connected neural network to derive predicted binding affinities.
	\item \textbf{WideDTA} \cite{widedta}: Based on DeepDTA, WideDTA introduces more chemical and biological text information to strengthen the representation of drugs and targets.
	\item \textbf{DeepCDA} \cite{m2}: The DeepCDA model is improved on the basis of the DeepDTA model, and its representation learning module combines convolutional neural networks and LSTM.
	\item \textbf{GraphDTA} \cite{m4}: The GraphDTA model uses molecular graphs and graph neural networks to learn the representation of drugs, and uses protein sequences and 1D convolutional neural networks to learn the representation of targets. The prediction module uses a multi-layer fully connected neural network.
	\item \textbf{DeepGLSTM} \cite{m5}: The DeepGLSTM model is improved on the basis of GraphDTA, which uses a bidirectional LSTM model as a target representation learning module to improve the target representation.
\end{itemize}

Table 4 shows the experimental results of the GraphCL-DTA model and all state-of-the-art models. It is obvious that with the support of graph contrastive learning and representation uniformity, the GraphCL-DTA model has achieved the best performance in all datasets and evaluation metrics. The metrics on the Davis dataset are 0.236, 0.894, 0.685. The metrics on the KIBA dataset are 0.129, 0.895, 0.805.

Firstly, the input of GraphDTA and DeepGLSTM is consistent with the GraphCL-DTA model. The metric values of GraphDTA on the Davis dataset are 0.241, 0.887, 0.679. The metric values on the KIBA dataset are 0.151, 0.883, and 0.687. The metric values of DeepGLSTM on the Davis dataset are 0.236, 0.893, and 0.677. The metric values on the KIBA dataset are 0.143, 0.890, and 0.780. Compared with GraphDTA, the average improvement of the GraphCL-DTA model on the two datasets is 1.2\% and 11\%. Compared with DeepGLSTM, the GraphCL-DTA model has an average improvement of 0.4\% and 4.5\% on the two datasets. Such improvement indicates that the graph contrastive learning and representation uniformity are of great help to the predictive ability of the model.

Since the GraphDTA and DeepGLSTM models only optimize the representation learning module, they do not utilize the structure of the molecular graph of the drug itself, which leads to its poor performance. The GraphCL-DTA model considers the molecular structure of the drug, and uses graph contrastive learning to mine the internal information of the structure, the performance of which is better as a result.

In addition, the DeepDTA, WideDTA and DeepCDA models that use the 1D structure of the drug as input perform worse than the GraphDTA and DeepGLSTM models that use the 2D structure of the drug as input. This shows that the amount of information contained in the molecular graph structure is greater than that of the SMLIES string.

The analysis and discussion of the above experimental results prove that the GraphCL-DTA model has better performance than the state-of-the-art models, thus illustrating the excellence of the model proposed in this work. There are two main reasons. Firstly, the graph contrastive learning framework that retains molecular semantic information reduces its dependence on supervised data, which helps learn the more essential and effective drug representation. Secondly, the loss function optimized for the uniformity of drug representation and target representation improves the quality of representation.

\section{Conclusion}

In this work, we design a graph contrastive learning framework for molecular graphs to learn drug representation, which preserves the semantic information of molecular graphs without data augmentation strategies. Through this graph contrastive framework, the more essential and effective drug representation can be learned without additional supervised data. In this way, the dependence on wet experiment data can be reduced, resulting in significant cost savings. In addition, we propose a loss function optimized for the uniformity of drug representation and target representation, which can be used to directly improve the quality of representation. Extensive experiments on two public datasets demonstrate the effectiveness of the graph contrastive learning framework and uniformity loss function. Moreover, the GraphCL-DTA model is verified to outperform the state-of-the-art model.
However, the GraphCL-DTA model only considers the features of atoms when extracting the drug representation, and does not take into account the features of the bonds between atoms, which loses the information contained in part of the molecular graph. Besides, the 1D CNN cannot capture the correlation information between protein sequences. In future work, we will develop GCN models that can consider both atomic and bond features in molecular graphs for learning more expressive drug representation. And more advanced sequence models will be used, such as transformer, to extract the representation of the target to ensure the better representation of the target.

\bibliography{reference.bib}

\end{document}